\documentclass[10pt,twocolumn,letterpaper]{article}

\usepackage{cvpr}
\usepackage{times}
\usepackage{epsfig}
\usepackage{graphicx}
\usepackage{amsmath}
\usepackage{amssymb}

\usepackage{bm}
\newcommand{\argmax}{\mathop{\rm argmax}\limits}
\usepackage{algorithm}
\usepackage{algorithmic}

\newcommand{\bvec}[1]{{\mbox{\boldmath $#1$}}}

\newcommand{\Tref}[1]{Table~\ref{#1}}
\newcommand{\Eref}[1]{Eq.~(\ref{#1})}
\newcommand{\Fref}[1]{Fig.~\ref{#1}}
\newcommand{\Aref}[1]{Algorithm~\ref{#1}}
\newcommand{\Sref}[1]{Section~\ref{#1}}
\newcommand{\APref}[1]{Appendix~\ref{#1}}

\usepackage[T1]{fontenc}
\usepackage{lmodern}

\usepackage[breaklinks=true,bookmarks=false]{hyperref}

\cvprfinalcopy 

\def\cvprPaperID{3107} 

\begin{document}

\title{Joint Optimization Framework for Learning with Noisy Labels}

\author{Daiki Tanaka\ \ \ \ Daiki Ikami\ \ \ \ Toshihiko Yamasaki\ \ \ \ Kiyoharu Aizawa\\
The University of Tokyo, Japan\\
{\tt\small \{tanaka, ikami, yamasaki, aizawa\}@hal.t.u-tokyo.ac.jp}
}

\maketitle

\begin{abstract}
  Deep neural networks (DNNs) trained on large-scale datasets have exhibited significant performance in image classification. Many large-scale datasets are collected from websites, however they tend to contain inaccurate labels that are termed as noisy labels. Training on such noisy labeled datasets causes performance degradation because DNNs easily overfit to noisy labels. To overcome this problem, we propose a joint optimization framework of learning DNN parameters and estimating true labels. Our framework can correct labels during training by alternating update of network parameters and labels. We conduct experiments on the noisy CIFAR-10 datasets and the Clothing1M dataset. The results indicate that our approach significantly outperforms other state-of-the-art methods.
\end{abstract}

\section{Introduction}\label{sec:intro}
DNNs trained on large-scale datasets have achieved impressive results on many classification problems. Generally, accurate labels are necessary to effectively train DNNs. However, many datasets are constructed by crawling images and labels from websites and often contain incorrect noisy labels (\eg, YFCC100M~\cite{thomee2015new}, Clothing1M~\cite{xiao2015learning}). This study addresses the following question: how can we effectively train DNNs on noisy labeled datasets without manually cleaning the data?

The prominent issue in training DNNs on noisy labeled datasets is that \textit{DNNs can learn or memorize, any training dataset}, and this implies that DNNs are subject to total overfitting on noisy data.

To address this problem, commonly used regularization techniques including dropout and early stopping are helpful. However, these methods do not guarantee optimization because they prevent the networks from reducing the training loss. Another method involves using prior knowledge, such as the confusion matrix between clean and noisy labels, which typically cannot be used in real settings.

\begin{figure}[t]
  \centering
  \scalebox{0.4}{
  \includegraphics{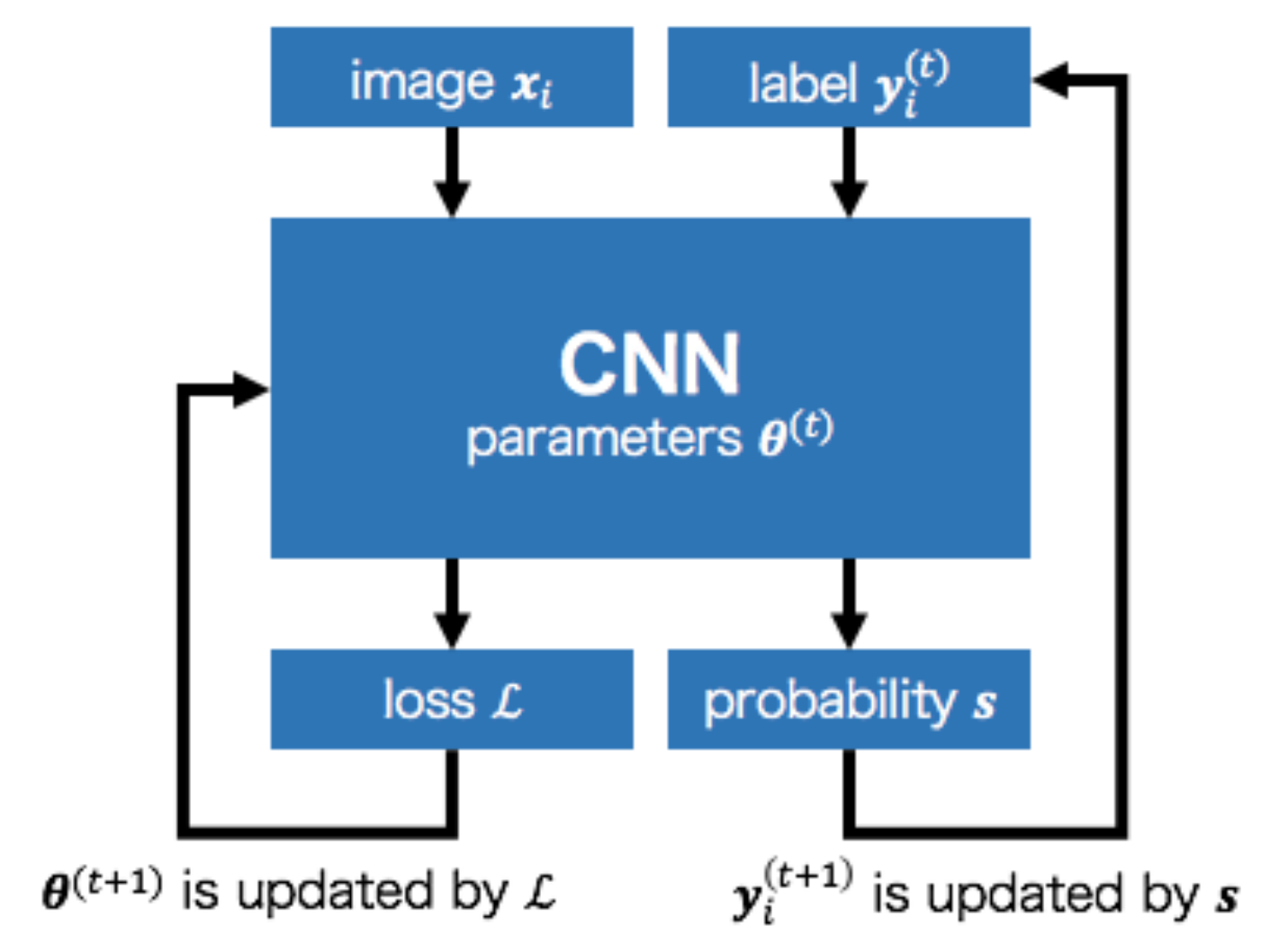}
  }
  \caption{The concept of our joint optimization framework. Noisy labels are reassigned to the probability output by CNNs. Network parameters and labels are alternatively updated for each epoch.}
  \label{fig:gainen}
  \vspace{-5mm}
\end{figure}

Consequently, we need a new framework of optimization. In this study, we propose an optimization framework for learning on a noisy labeled dataset. We propose optimizing the labels themselves as opposed to treating the noisy labels as fixed. The joint optimization of network parameters and the noisy labels corrects inaccurate labels and simultaneously improves the performance of the classifier. \Fref{fig:gainen} shows the concept of our proposal. The main contributions are as follows.

\begin{enumerate}
	\item {We propose a joint optimization framework for learning on noisy labeled datasets. Our optimization problem has two optimization network parameters and class labels that are optimized by an alternating strategy.}
	\vspace{-2mm}
	\item{We observe that a DNN trained on noisy labeled datasets does not memorize noisy labels and maintains high performance for clean data under a high learning rate. This reinforces the findings of Arpit \etal~\cite{arpit2017closer} that suggest that DNNs first learns simple patterns and subsequently memorize noisy data.}
	\vspace{-2mm}
	\item{We evaluate the performance on synthetic and real noisy datasets. We demonstrate state-of-the-art performance on the noisy CIFAR-10 dataset and a comparable performance on the Clothing1M dataset~\cite{xiao2015learning}.}
\end{enumerate}

\section{Related Works}
\subsection{Generalization abilities of DNNs}
\label{sec:generalization}
Recently, generalization and memorization abilities of neural networks have attracted increasing attention. Specifically, we focus on the ability of learning labels. Zhang \etal showed that DNNs can learn any training dataset even if the training labels are completely random~\cite{zhang2016understanding}. This leads to two problems.  First, the performance of a DNN decreases when it is trained on a noisy dataset and completely learns noisy labels. Second, it is difficult to learn which label is noisy given the perfect learning ability. To the best of our knowledge, most studies on deep learning with respect to noisy labels do not focus on the aforementioned problems that are caused by the memorization ability of DNNs. This study involves addressing these two problems to improve the classification accuracy by preventing completely learning for noisy labels.

\subsection{Learning on noisy labeled datasets}
We briefly review existing studies on learning on noisy labeled datasets.

\vspace{2mm}\noindent\textbf{Regularization:}
Regularization is an efficient method to deal with the issue of DNNs easily fitting noisy labels, as described in~\Sref{sec:generalization}. Arpit \etal showed the performances of DNNs trained on noisy labeled datasets with several regularizations~\cite{arpit2017closer} including weight decay, dropout, and adversarial training~\cite{goodfellow2014explaining}.
Zhang \etal used a mixup~\cite{zhang2017mixup} involving the utilization of a linear combination of images and labels for training.

These techniques improve performance on clean labels. However, these methods do not explicitly deal with noisy labels, and therefore long-time training leads to performance degradation as follows: the performance of the last epoch is generally worse than that of the best epoch~\cite{zhang2017mixup}. Furthermore, it is not possible to use the training loss on the noisy labeled dataset as a measure of performance on clean labels. Therefore, training-loss based early stopping does not work well.

\vspace{2mm}\noindent\textbf{Noise transition matrix:}
Let $l$ and $l^{GT}$ be the noisy and true labels. We define the noise transition matrix $T$ by $t_{ij} = p(l=j|l^{GT}=i)$. Then, we can use $T$ to modify the cross entropy loss as follows:
\begin{equation}\label{eq:NoisematrixCE}
\mathcal{L}(\bm{\theta},Y,X) = -\frac{1}{n}\sum_{i=1}^n\log\left(\bm{y}_i^{\mathrm{T}}T\bm{s}(\bm{\theta},\bm{x}_i)\right).
\end{equation}
This formulation was used in many studies~\cite{sukhbaatar2014training,jindal2016learning,patrini2016making}.
In deep learning, some studies presuppose the ground-truth noise transition matrix $T$~\cite{patrini2016making,vahdat2017toward} and achieve the state-of-the-art performance in the noisy CIFAR-10 dataset. Other studies estimate $T$ from noisy data. Specifically, $T$ is modeled by a fully connected layer and is trained in an end-to-end manner~\cite{sukhbaatar2014training,jindal2016learning}.
However, these methods do not carefully focus on the memorization ability of DNNs. Patrini \etal proposed an estimation method for $T$~\cite{patrini2016making}; however, its performance is slightly worse than that obtained with the true $T$.

\vspace{2mm}\noindent\textbf{Robust loss function:}
A few studies achieve noise-robust classification by using a noise-tolerant loss functions, such as ramp loss~\cite{brooks2011support} and unhinged loss~\cite{van2015learning}. For further details please refer to~\cite{ghosh2015making}. In deep learning, Ghosh \etal used mean square error and mean absolute error~\cite{ghosh2017robust} for noise-tolerant loss functions. It should be noted that they do not consider the problem that DNNs can learn arbitrary labels.

\vspace{2mm}\noindent\textbf{Other approaches using deep learning:}
Reed \etal used a bootstrapping scheme to handle noisy labels~\cite{reed2014training}. Our method is similar to this study.
Xiao \etal constructed a noise model with multiple noise types and trained two networks: an image classifier and a noise type classifier~\cite{xiao2015learning}. It should be noted that this method requires a low amount of accurately labeled datasets.

\subsection{Self-training and pseudo-labeling}
Pseudo-labeling~\cite{zhu2006semi,haffari2012analysis,lee2013pseudo} is a type of self-training that is generally used in semi-supervised learning with few labeled data and many unlabeled data. In this technique, pseudo-labels are initially assigned to unlabeled data by predictions of a model trained on a clean dataset. Subsequently, the algorithm repeats retraining the model on both labeled and unlabeled data and updating pseudo-labels.

In semi-supervised learning, we know which data is labeled or not and only need to assign pseudo-labels to only unlabeled data. However, with respect to learning on noisy labeled data, it is necessary to treat all data equally because we do not know which data is noisy. Reed \etal proposed a self-training scheme~\cite{reed2014training} for training a DNN on noisy labeled data. Their approach is similar to that proposed in this study. However, they use original noisy labels for learning until the end of training, and thus the performance is degraded by the remaining effects of noisy labels for a high noise rate~\cite{reed2014training,jindal2016learning}. Conversely, we completely replace all labels by pseudo-labels and use the same for training.

\section{Notation and Problem Statements}
In this study, column vectors and matrices are denoted in bold (\eg $\bm{x}$) and capitals (\eg $X$), respectively. Specifically, $\bvec{1}$ is a vector of all-ones.
We define hard-label spaces $\mathcal{H}=\{\bm{y}:\bm{y}\in\{0,1\}^c, \bm{1}^\top\bm{y}=1\}$ and soft-label spaces $\mathcal{S}=\{\bm{y}:\bm{y}\in[0,1]^c, \bm{1}^\top\bm{y}=1\}$.

In supervised $c$-class classification problem setting, we have a set of $n$ training images $X=[\bm{x}_1,\dots,\bm{x}_n]$
with ground-truth labels $Y^{GT}=[\bm{y}^{GT}_1,\dots,\bm{y}^{GT}_n]=\mathbb{R}^{c\times n}$,
where $\bm{y}^{GT}_i\in\mathcal{H}$ is a one-hot vector representation of the true class label. The objective function is an empirical risk, such as the cross entropy, as follows:
\begin{equation}
\mathcal{L}=-\frac{1}{n}\sum_{i=1}^n\sum_{j=1}^cy^{GT}_{ij}\log s_j(\bm{\theta},\bm{x}_i),
\label{eq:supervisedLearning}
\end{equation}
where $\bm{\theta}$ denotes the set of network parameters and $\bm{s}$ denotes the output of the final layer, namely $c$-class softmax layer, of the network.

If a clean training dataset is present, then the network parameters $\bm{\theta}$ are learned by optimizing \Eref{eq:supervisedLearning} by using a gradient descent method. However, in this study, we consider the classification problem with noisy labels as follows: Let $\bm{y}_i$ be the noisy label, and only the noisy training label set $Y=[\bm{y}_1,\cdots,\bm{y}_n]$ is given. The task involves training CNNs to predict true labels. In the next section, we describe the proposed method for training on noisy labels.

\section{Classification with Label Optimization}
In this section, we present our proposed training method with noisy labels.
Generally, with respect to supervised learning on clean labels, the optimization problem is formulated as follows:
\begin{equation}\label{eq:usual}
\min_{\bm{\theta}}\mathcal{L}(\bm{\theta}|X,Y),
\end{equation}
where $\mathcal{L}$ denotes a loss function such as the cross entropy loss \Eref{eq:supervisedLearning}. \Eref{eq:usual} works well on clean labels. However, if we train the network by \Eref{eq:usual} on noisy labels, its performance decreases.

As we will describe in \Sref{sec:prelim}, we experimentally found that a high learning rate suppresses the memorization ability of a DNN and prevents it from completely fitting to labels. Thus, we assume that a network trained with a high learning rate will have more difficulty fitting to noisy labels. In other words, the loss \Eref{eq:usual} is high for noisy labels and low for clean labels. Given this assumption, we obtain clean labels by updating labels in the direction to decrease \Eref{eq:usual}. Therefore, we formulate the problem as the joint optimization of network parameters and labels as follows:
\begin{equation}\label{eq:modified}
\min_{\bm{\theta},Y}\mathcal{L}(\bm{\theta},Y|X).
\end{equation}
The concept of our proposal is shown in \Fref{fig:gainen}.

Our proposed loss function $\mathcal{L}(\bm{\theta},Y|X)$ is constructed by three terms as follows:
\begin{equation}\label{eq:all}
  \mathcal{L}(\bm{\theta},Y|X)=\mathcal{L}_c(\bm{\theta},Y|X)+\alpha\mathcal{L}_{p}(\bm{\theta}|X)+\beta\mathcal{L}_{e}(\bm{\theta}|X),
\end{equation}
where $\mathcal{L}_c(\bm{\theta},Y|X)$, $\mathcal{L}_{p}(\bm{\theta}|X)$, $\mathcal{L}_{e}(\bm{\theta}|X)$ denote the classification loss and two regularization losses, respectively, and $\alpha$ and $\beta$ denote hyper parameters.
In this study, we use the Kullback-Leibler (KL)-divergence for $\mathcal{L}_c(\bm{\theta},Y|X)$ as follows:
\begin{equation}\label{eq:KLdivergence}
\mathcal{L}_c(\bm{\theta},Y|X) = \frac{1}{n}\sum_{i=1}^n D_{KL}(\bm{y}_i||\bm{s}(\bm{\theta},\bm{x}_i)),
\end{equation}
\begin{equation}\label{eq:KLdivergence2}
D_{KL}(\bm{y}_i||\bm{s}(\bm{\theta},\bm{x}_i))=\sum_{j=1}^c y_{ij}\log\left(\frac{y_{ij}}{s_j}(\bm{\theta},\bm{x}_i)\right).
\end{equation}
In the following subsections, we first describe an alternating optimization method to solve this problem, and we then describe the definition of $\mathcal{L}_{p}$, $\mathcal{L}_{e}$.

\subsection{Alternating Optimization}\label{sec:alt}
In our proposed learning framework, network parameters $\bm{\theta}$ and class labels $Y$ are alternatively updated as shown in \Aref{alg1}.
We will describe the update rules of $\bm{\theta}$ and $Y$.

\addtolength\textfloatsep{-5mm}
\begin{algorithm}[t]
	\caption{Alternating Optimization}
	\label{alg1}
	\begin{algorithmic}
		\FOR{$t\leftarrow 1$ to $num\_epochs$}
		\STATE $\text{update }\bm{\theta}^{(t+1)}\text{ by SGD on }\mathcal{L}(\bm{\theta}^{(t)},Y^{(t)}|X)$
		\STATE $\text{update }Y^{(t+1)}\text{ by \Eref{eq:label} (hard-label)}$
    \STATE $\text{\quad\quad\quad\quad\quad\quad\ \  or \Eref{eq:soft} (soft-label)}$
		\ENDFOR
	\end{algorithmic}
\end{algorithm}

\vspace{2mm}\noindent\textbf{Updating $\bm{\theta}$ with fixed $Y$:}
All terms in the optimization problem \Eref{eq:all} are sub-differentiable with respect to $\bm{\theta}$. Therefore, we update $\bm{\theta}$ by the stochastic gradient descent (SGD) on the loss function \Eref{eq:all}.

\vspace{2mm}\noindent\textbf{Updating $Y$ with fixed $\bm{\theta}$:}
In contrast to other methods, we update and optimize the labels that we perform the training on.
With respect to updating $Y$, it is only necessary to consider the classification loss $\mathcal{L}_c(\bm{\theta},Y|X)$ from \Eref{eq:all} with fixed $\bm{\theta}$. The optimization problem \Eref{eq:KLdivergence} on $Y$ is separated for each $\bm{y}_i$.

As a method of optimizing labels, two methods can be considered: the hard-label method and the soft-label method.
In the case of the hard-label method, $Y\in\mathcal{H}$ is updated as follows:
\begin{equation}\label{eq:label}
y_{ij} =
\begin{cases}
1\quad\text{if}\quad j = \argmax_{j'}s_{j'}(\bm{\theta},\bm{x}_i)\\
0\quad\text{otherwise}
\end{cases}.
\end{equation}
In the case of the soft-label method, the KL-divergence from $\bm{s}(\bm{\theta},\bm{x}_i)$ to $\bm{y}_i$ is minimized when $\bm{y}_i=\bm{s}$, and thus the update rule for $Y\in\mathcal{S}$ is as follows:
\begin{equation}\label{eq:soft}
\bm{y}_i=\bm{s}(\bm{\theta},\bm{x}_i).
\end{equation}
As we will describe in \Sref{sec:vs}, we experimentally determined that the performance of the soft-label method exceeded that of the hard-label method. Thus, we applied soft-labels to all experiments if not otherwise specified.

\addtolength\textfloatsep{5mm}
\subsection{Regularization Terms}\label{sec:regterm}
We describe definitions and roles of two regularization losses of $\mathcal{L}_p(\bm{\theta}|X)$ and $\mathcal{L}_e(\bm{\theta}|X)$.

\vspace{2mm}\noindent\textbf{Regularization loss $\mathcal{L}_p$:}
The regularization loss $\mathcal{L}_p(\bm{\theta}|X)$ is required to prevent the assignment of all labels to a single class: In the case of minimizing only \Eref{eq:KLdivergence}, we obtain a trivial global optimal solution with a network that always predicts constant one-hot $\hat{\bm{y}}\in\mathcal{H}$ and each label $\bm{y}_i=\hat{\bm{y}}$ for any image $\bm{x}_i$.
To overcome this problem, we introduce a prior probability distribution $\bm{p}$, which is a distribution of classes among all training data. If the prior distribution of classes is known, then the updated labels should follow the same. Therefore, we introduce the KL-divergence from $\bar{\bm{s}}(\bm{\theta},X)$ to $\bm{p}$ as a cost function as follows:
\begin{equation}\label{eq:kl}
\begin{split}
\mathcal{L}_{p}=\sum_{j=1}^cp_j\log\frac{p_j}{\bar{s}_j(\bm{\theta},X)}
\end{split}
\end{equation}
This approach follows~\cite{hu2017learning}. The mean probability $\bar{\bm{s}}(\bm{\theta},X)$ in the training data is approximated by performing a calculation for each mini-batch $\mathcal{B}$ as \Eref{eq:batch}.
\begin{equation}\label{eq:batch}
\bar{\bm{s}}(\bm{\theta},X)
=\frac{1}{n}\sum_{i=1}^n\bm{s}(\bm{\theta},\bm{x}_i)\approx\frac{1}{|\mathcal{B}|}\sum_{\bm{x}\in\mathcal{B}}\bm{s}(\bm{\theta},\bm{x})
\end{equation}
This approximation cannot treat a large number of classes and extreme imbalanced classes, however it works well in the experiments on the noisy CIFAR-10 dataset and the Clothing1M dataset.

\vspace{2mm}\noindent\textbf{Regularization loss $\mathcal{L}_e$:}
The term $\mathcal{L}_e$ is required for the training loss when we use the soft-label. We consider the case of \Eref{eq:all} with $\alpha=\beta=0$. In this case, when $Y$ is updated by \Eref{eq:soft}, both $\bm{\theta}$ and $Y$ are stuck in local optima and the learning process does not proceed. To overcome this problem, we introduce an entropy term to concentrate the probability distribution of each soft-label to a single class as follows:
\begin{equation}\label{eq:ent}
\mathcal{L}_{e} = -\frac{1}{n}\sum_{i=1}^n\sum_{j=1}^cs_j(\bm{\theta},\bm{x}_i)\log s_j(\bm{\theta},\bm{x}_i).
\end{equation}

\subsection{Additional Details}
Our method has two steps for training on noisy labels. In the first step, we obtain clean labels by updating labels as described in \Sref{sec:alt}. In the second step, we initialize the network parameters and train the network by usual supervised learning with the labels obtained in the first step.

\section{Experiments}
\subsection{Datasets}
\noindent\textbf{CIFAR-10:}
We use the CIFAR-10 dataset~\cite{krizhevsky2009learning} and retain 10\% of the training data for validation. Subsequently, we define three types of the training data, namely Symmetric Noise CIFAR-10 (SN-CIFAR), Asymmetric Noise CIFAR-10 (AN-CIFAR), and Pseudo Label CIFAR-10 (PL-CIFAR).

In SN-CIFAR, we inject the symmetric label noise. Symmetric label noise is as follows:
\begin{equation}
\bm{y}_i = \begin{cases}
\bm{y}^{GT}_i \mbox{ with the probability of }1-r \\
\mbox{random one-hot vector with the probability of }r
\end{cases}.
\end{equation}

In AN-CIFAR, we inject the asymmetric label noise. The asymmetric label noise is discussed in~\cite{patrini2016making}. The rationale involves mimicking a part of the structure of real mistakes for similar classes: TRUCK $\rightarrow$ AUTOMOBILE, BIRD $\rightarrow$ AIRPLANE, DEER $\rightarrow$ HORSE, CAT $\leftrightarrow$ DOG. Transitions are parameterized by $r\in[0,1]$ such that the probabilities of ground-truth and inaccurate class correspond to $1-r$ and $r$, respectively.

In PL-CIFAR, pseudo labels are assigned to unlabeled training data. Pseudo labels are generated by applying k-means++~\cite{arthur2007k} to features that are outputs of pool5 layer of ResNet-50~\cite{he2016deep} pre-trained on ImageNet. This setting is motivated by transfer learning. The overall accuracy of the pseudo labels is 62.50\%.

\vspace{2mm}\noindent\textbf{Clothing1M:}
We use the Clothing1M dataset~\cite{xiao2015learning} to examine the performance of our method in a real setting.
The Clothing1M dataset contains 1 million images of clothing obtained from several online shopping websites that are classified into the following 14 classes: T-shirt, Shirt, Knitwear, Chiffon, Sweater, Hoodie, Windbreaker, Jacket, Down Coat, Suit, Shawl, Dress, Vest, and Underwear.
The labels are generated by using surrounding texts of the images that are provided by the sellers, and therefore contain many errors.
In~\cite{xiao2015learning}, it is reported that the overall accuracy of the noisy labels is 61.54\%, and some pairs of classes are often confused with each other (\eg, Knitwear and Sweater).
The Clothing1M dataset also contains $50k$, $14k$ and $10k$ of clean data for training, validation, and testing, respectively although we do not use the $50k$ clean training data.

\subsection{Implementation}
We implemented all the models with the deep learning framework Chainer v2.1.0~\cite{tokui2015chainer}.

\vspace{2mm}\noindent\textbf{CIFAR-10:}
Training on SN-CIFAR, AN-CIFAR and PL-CIFAR, we used the network based on PreAct ResNet-32~\cite{he2016identity} as detailed in \APref{sec:arch}.
With respect to preprocessing, we performed mean subtraction and data augmentation by horizontal random flip and 32$\times$32 random crops after padding with 4 pixels on each side. We used SGD with a momentum of 0.9, a weight decay of $10^{-4}$, and batch size of 128.

In the first step of our method, we trained the network for 200 epochs and began updating labels from the 70th epoch. We determined the values of a learning rate and the hyper parameters ($\alpha$, $\beta$ in \Eref{eq:all}) for SN-CIFAR, AN-CIFAR, and PL-CIFAR respectively based on the validation accuracy. The details are described in each experimental section. As we will describe in \Sref{sec:vs}, soft-labels performed better than hard-labels, and thus we applied soft-labels to all the experiments in \Sref{sec:uni}, \Sref{sec:asym}, and \Sref{sec:pl}.
In this case, the prior distribution $p$ is uniform distribution because each class has the same number of images in the CIFAR-10 dataset.
While updating the noisy label $\bm{y}_i$ by the probability $\bm{s}$, we used the average output probability of the network of the past 10 epochs as $\bm{s}$. We experimentally determined that this averaging technique is useful in preventing inaccurate updates since it has a similar effect to ensemble.

In the second step of our method, we trained the network for 120 epochs on the labels obtained in the first step. We began training with a learning rate of 0.2 and divided it by 10 after 40 and 80 epochs. We used only $\mathcal{L}_c$ for the training loss in this step.

\vspace{2mm}\noindent\textbf{Clothing1M:}
Training on the Clothing1M dataset, we used ResNet-50 pre-trained on ImageNet to align experimental condition with \cite{patrini2016making}.
For preprocessing, we resized the images to $256\times256$, performed mean subtraction, and cropped the middle $224\times224$. We used SGD with a momentum of 0.9, a weight decay of $10^{-3}$, and batch size of 32.

In the first step of our method, we trained the network for 10 epochs and began updating labels from the 1st epoch. We used a learning rate of $8\times10^{-4}$, and used 2.4 for $\alpha$ and 0.8 for $\beta$. While updating the noisy label $\bm{y}_i$ by the probability $\bm{s}$, we used the average output probability of the network of all the past epochs as $\bm{s}$. We applied soft-labels to the experiment in \Sref{sec:cloth}.

In the second step of our method, we trained the network for 10 epochs on the labels obtained in the first step. We began training with a learning rate of $5\times10^{-4}$ and divided it by 10 after 5 epochs.

\subsection{Generalization and Memorization}\label{sec:prelim}
\begin{figure}[tb]
  \vspace{-6mm}
  \centering
  \scalebox{0.5}{
  \includegraphics{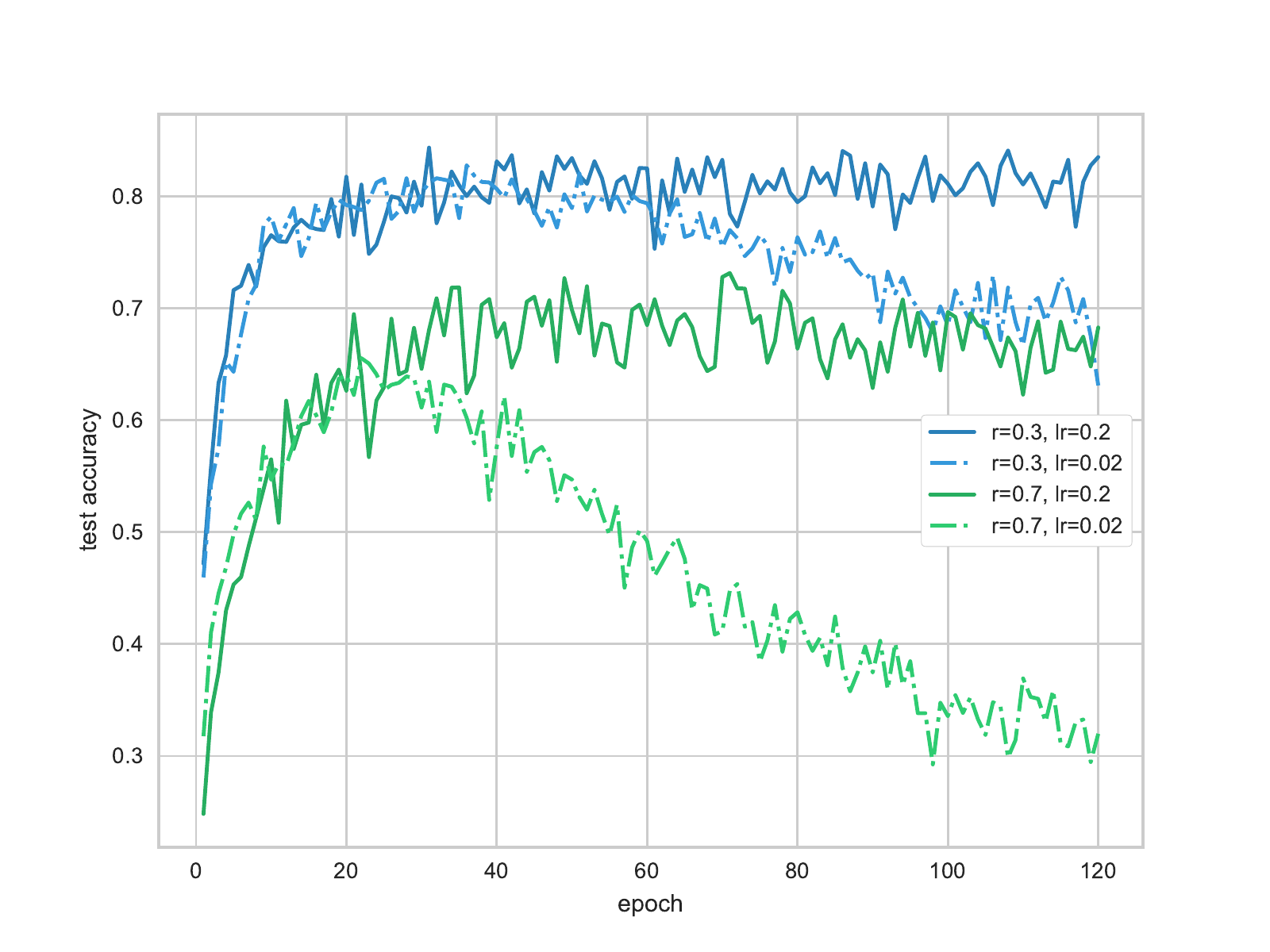}
  }
  \caption{The test accuracy curve with different learning rates. The test accuracy gradually decreases when the learning rate is low (lr=0.02). Conversely, the test accuracy remains high at the end of training when the learning rate is high (lr=0.2).}
  \label{fig:acc}
  \vspace{-3mm}
\end{figure}
\begin{figure}[tb]
  \centering
  \scalebox{0.5}{
  \includegraphics{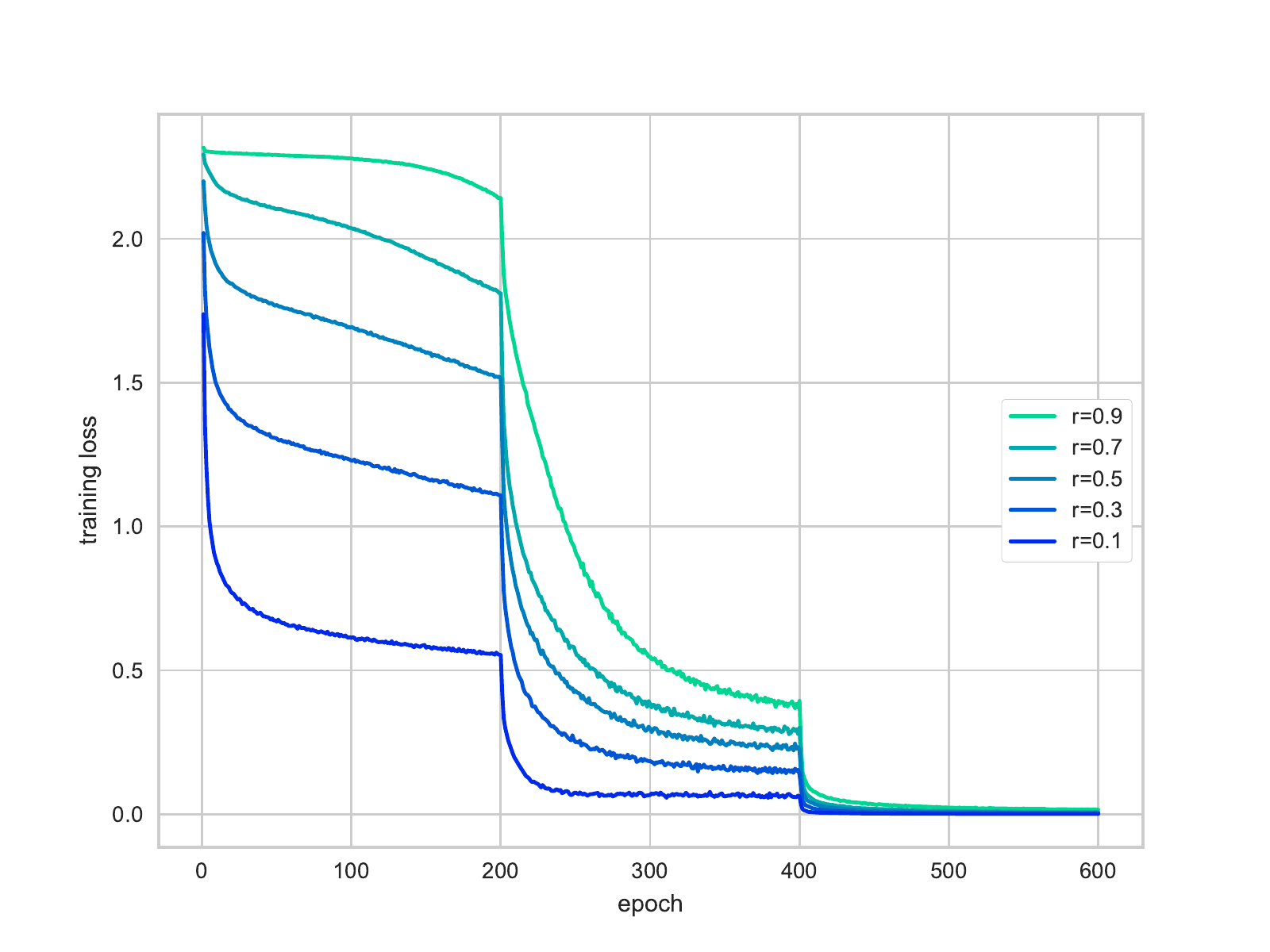}
  }
  \caption{The training loss curve with different noise rates. At the end of training with a low learning rate, the value of the training loss is close to 0 even if the error rate is 0.9. In contrast, in the early phase of training with a high learning rate, an increase in the noise rate increases the training loss.}
  \label{fig:loss}
  \vspace{-3mm}
\end{figure}

To examine the effect of the learning rate (lr) and the noise rate ($r$) on the training loss and the test accuracy, we trained the network on SN-CIFAR with only the cross entropy loss.

\Fref{fig:acc} shows the test accuracy curve with different learning rates. We trained the network for 120 epochs with a learning rates of 0.2 or 0.02. In the case of the low learning rate (lr=0.02), the test accuracy was high at the early phase of training and then gradually decreased because the network fitted the noisy labels. This is the same result reported in~\cite{arpit2017closer}. Conversely, in the case of the high learning rate (lr=0.2), the network exhibited high test accuracies during training. This means that a high learning rate prevents the network from memorizing and fitting the noisy labels.

\Fref{fig:loss} shows how the manner in which training loss declines during training. We trained the network for 600 epochs. We commenced training with a learning rate of 0.2 and divided it by 10 after 200 and 400 epochs. At the end of training, our model fit the noisy labels even if the noise rate was high (for \eg, $r=0.9$). However, with respect to training with a high learning rate, the training loss clearly increases when the noise rate increases. This indicates that it is possible to optimize the labels towards lowering the training loss when the learning rate is high.

\subsection{Hard-Label vs. Soft-Label}\label{sec:vs}
\begin{figure}[t]
  \vspace{-5mm}
  \centering
  \scalebox{0.5}{
  \includegraphics{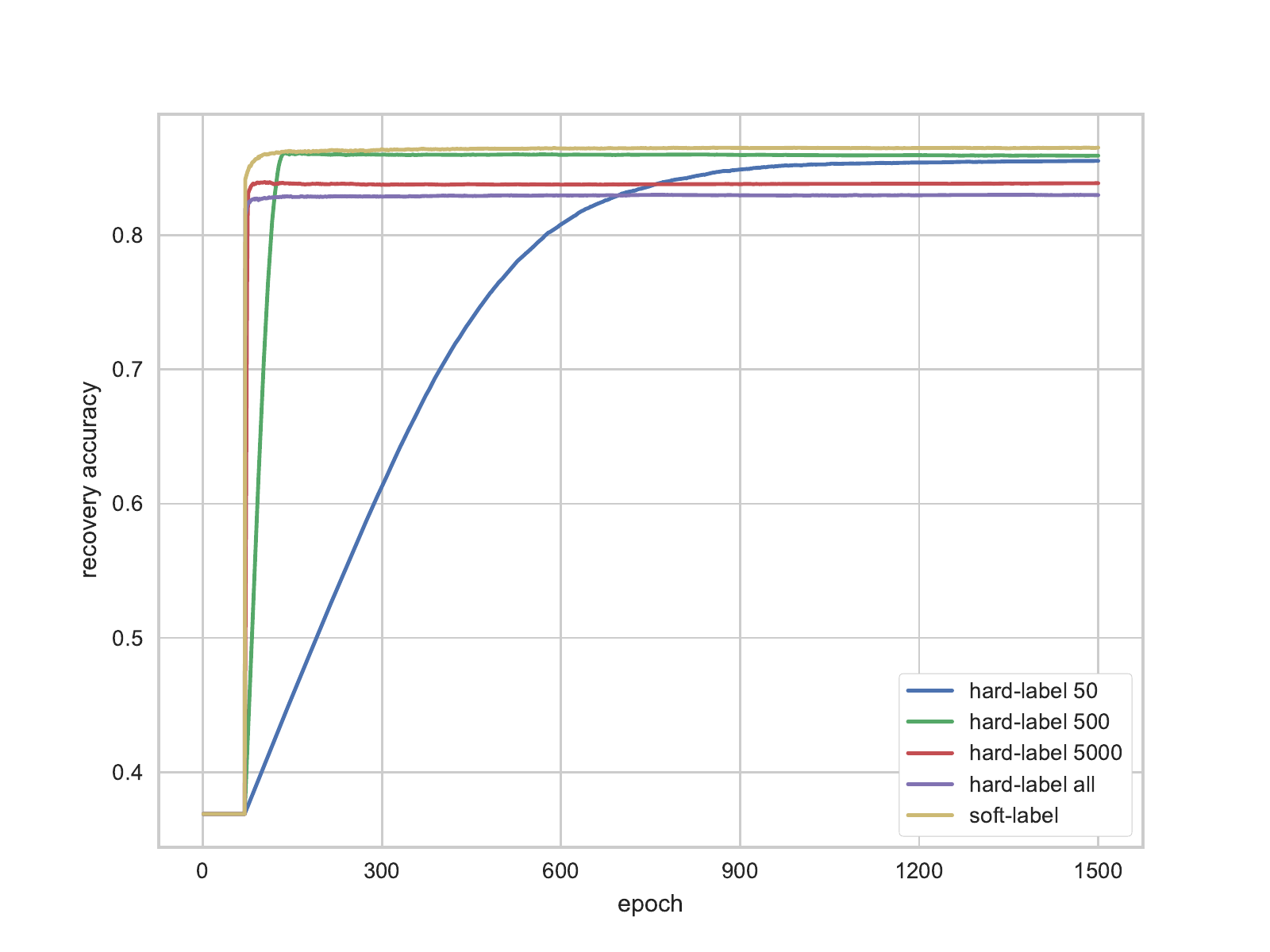}
  }
  \caption{Comparison between the soft-label and the hard-label methods, showing the recovery accuracy. The soft-label method achieves faster convergence than the hard-label methods, and performs the best recovery accuracy.}
  \label{fig:hard}
  \vspace{-5mm}
\end{figure}

To prove the effectiveness of the soft-label, we trained the network on SN-CIFAR (noise rate $r=0.7$) for 1500 epochs by using the first step of our method. We compare the hard-label methods and the soft-label method. For the hard-labels methods, we update top 50, 500, 5000, or all labels whose current labels are most different from the predicted classes to the predicted hard-labels every epoch. For the soft-label method, we update all labels to the predicted soft-labels every epoch.
In \Fref{fig:hard}, we show the recovery accuracy, which is defined as the accuracy of the reassigned labels, in the first step of our method. The soft-label method achieves faster convergence and better recovery accuracy than any hard-label methods.

Subsequently, by using the second step of our method, we performed training on the labels obtained in the first step. In the hard-label methods, updating 500 labels every epoch is optimal and the test accuracy is 85.7\%. Conversely, the test accuracy of the soft-label method is 86.0\%. It shows that though the recovery accuracy of the soft-label method obtained in the first step is 86.0\%, which is approximately equal to 85.9\% of the hard-label method (updating 500 labels every epoch), the test accuracy is improved by 0.3\%. The reason why the soft-label method performed better is considered as that soft-labels contain the probabilities of each class in themselves.
Soft-labels reflect confidences of the trained network unlike hard-labels, which are assigned by ignoring confidences. Our results indicate that confidences are important in the case of training on noisy labels.

\subsection{Experiment on SN-CIFAR}\label{sec:uni}
\begin{table*}[t]
  \vspace{-3mm}
  \caption{Test and recovery accuracy of different baselines on the CIFAR-10 dataset with symmetric noise. We report the average score of 5 trials.}
  \vspace{-2mm}
  \begin{center}
  \scalebox{0.9}{
  \begin{tabular}{ccc|cccccc|cccccc}\hline
    \multicolumn{1}{c|}{\#}&method&&\multicolumn{6}{c}{Test Accuracy (\%)}&\multicolumn{6}{|c}{Recovery Accuracy (\%)}\\ \hline
    \multicolumn{3}{c|}{noise rate (\%)}&0&10&30&50&70&90&0&10&30&50&70&90\\ \hline\hline
    \multicolumn{1}{c|}{1}&Cross Entropy Loss &\textit{best}& 93.5 & 91.0 & 88.4 & 85.0 & 78.4 & 41.1 &\textbf{100.0}& 96.4 & 92.7 & 88.2 & 80.1 & 41.4\\
    \multicolumn{1}{c|}{}&&\textit{last}& 93.4 & 87.0 & 72.2 & 55.3 & 36.6 & 20.4 &\textbf{100.0}& 91.1 & 74.6 & 57.6 & 39.6 & 21.7\\ \hline
    \multicolumn{1}{c|}{2}&Our Method &\textit{best}& 93.4 & 92.7 & 91.4 & 89.6 & 85.9 & 58.0 &\textbf{100.0}& 97.9 &\textbf{95.1}& 91.7 & 86.3 & 58.2\\
    \multicolumn{1}{c|}{}&&\textit{last}&\textbf{93.6}&\textbf{92.9}&\textbf{91.5}&\textbf{89.8}&\textbf{86.0}&\textbf{58.3}& 99.9 &\textbf{98.1}&\textbf{95.1}&\textbf{91.8}&\textbf{86.4}&\textbf{58.3}\\ \hline
  \end{tabular}
  }
  \end{center}
  \label{tab:uni}
  \vspace{-5mm}
\end{table*}

To evaluate the performance of our method on synthesized noisy labels, we trained the network on SN-CIFAR (the noise rate $r=0.0,0.1,0.3,0.5,0.7,0.9$) by using our method.
In the first step of our method, we used the optimal learning rate, $\alpha$ and $\beta$ for each noise rate based on the validation accuracy as detailed in \APref{sec:hyper}.
As a comparison, we also trained on initial noisy labels in the same manner as the second step of our method.

The results are reported in \Tref{tab:uni}. As shown in \Tref{tab:uni}, \textit{best} denotes the scores of the epoch where the validation accuracy is optimal, and \textit{last} denotes the scores at the end of training.
The recovery accuracy for our method is defined as the accuracy of the reassigned labels.
Conversely, other methods do not reassign the noisy labels, and thus the recovery accuracy is reported as the prediction accuracy on the ground-truth labels of the noisy training data.

Our method achieves overall better test accuracy and recovery accuracy on SN-CIFAR.
When training was performed on initial noisy labels, the test accuracy decreases after approximately the 40th epoch (when we divided the learning rate by 10). This indicates that lowering the learning rate assists the network in fitting the noisy labels as described in \Sref{sec:prelim}. Conversely, when we trained on the labels optimized by our method, the test accuracy was high until the end of training. This is the important effects of our joint optimization.

\subsection{Experiment on AN-CIFAR}\label{sec:asym}
\begin{table*}[t]
  \caption{Test and recovery accuracy of different baselines on the CIFAR-10 dataset with asymmetric noise. We report the average score of 5 trials. \#2, \#3 are the results by our implementation.}
  \vspace{-2mm}
  \begin{center}
  \scalebox{0.9}{
  \begin{tabular}{ccc|ccccc|ccccc}\hline
    \multicolumn{1}{c|}{\#} & method &&\multicolumn{5}{c}{Test Accuracy (\%)}&\multicolumn{5}{|c}{Recovery Accuracy (\%)}\\ \hline
    \multicolumn{3}{c|}{noise rate (\%)}&10&20&30&40&50&10&20&30&40&50\\ \hline\hline
    \multicolumn{1}{c|}{1} & Cross Entropy Loss &\textit{best}& 91.8 & 90.8 & 90.0 & 87.1 & 77.3 & 97.2 & 95.8 & 94.3 & 91.0 & 80.5\\
    \multicolumn{1}{c|}{}&&\textit{last}& 89.8 & 85.4 & 81.0 & 75.7 & 70.5 & 95.0 & 90.2 & 85.3 & 80.2 & 75.2\\ \hline
    \multicolumn{1}{c|}{2} & Forward~\cite{patrini2016making} &\textit{best}& 92.4 & 91.4 & 91.0 & 90.3 & 83.8 & 97.7 & 96.7 & 95.9 & 94.7 & 88.0\\
    \multicolumn{1}{c|}{}&&\textit{last}& 91.7 & 89.7 & 88.0 & 86.4 & 80.9 & 97.9 & 95.8 & 93.6 & 91.5 & 85.5\\ \hline
    \multicolumn{1}{c|}{3} & CNN-CRF~\cite{vahdat2017toward} &\textit{best}& 92.0 & 91.5 & 90.7 & 89.5 & 84.0 & 97.4 & 96.5 & 95.3 & 93.7 & 88.1\\
    \multicolumn{1}{c|}{}&&\textit{last}& 90.3 & 86.6 & 83.6 & 79.7 & 76.4 & 95.1 & 90.5 & 86.4 & 82.1 & 78.7\\ \hline
    \multicolumn{1}{c|}{4} & Our Method &\textit{best}&\textbf{93.2}& 92.7 &\textbf{92.4}& 91.5 & 84.6 &\textbf{98.3}&\textbf{97.2}&\textbf{96.3}&\textbf{95.2}&\textbf{88.3}\\
    \multicolumn{1}{c|}{}&&\textit{last}&\textbf{93.2}&\textbf{92.8}&\textbf{92.4}&\textbf{91.7}&\textbf{84.7}& 98.1 & 97.1 &\textbf{96.3}&\textbf{95.2}& 88.1\\ \hline
  \end{tabular}
  }
  \end{center}
  \label{tab:asym}
  \vspace{-7mm}
\end{table*}

To evaluate the performance of our method in the settings in~\cite{patrini2016making}, we trained the network on AN-CIFAR (the noise rate $r=0.1,0.2,0.3,0.4,0.5$) by using our method.
In the first step of our method, we used a learning rate of 0.03 and used 0.8 for $\alpha$ and 0.4 for $\beta$, respectively for all the noise rates.
As a comparison, we also performed training on initial noisy labels in the same manner as the second step of our method with the cross entropy loss or the forward corrected loss~\cite{patrini2016making}.

The results of our experiments are shown in \Tref{tab:asym}.
The forward corrected loss~\cite{patrini2016making} and the CNN-CRF model~\cite{vahdat2017toward} require the ground-truth noise transition matrix. Conversely, we need only the prior distribution $p$, and thus our condition is more general than that of~\cite{patrini2016making, vahdat2017toward}.

Our method achieves significantly better test accuracy and recovery accuracy on AN-CIFAR.
However, only when the noise rate is 50\%, there is no significant improvement in accuracy when compared with other noise rates. Since we generated label noise to exchange CAT and DOG classes, it is impossible to accurately determine the class for CAT and DOG when the noise rate is 50\%.

In a manner similar to \Sref{sec:uni}, when training is performed on initial noisy labels, the test accuracy decreases due to the network fitting noisy labels with a low learning rate. This trend is also observed if we use the forward corrected loss~\cite{patrini2016making}, while the test accuracy does not decrease and remains high in our method.

\subsection{Experiment on PL-CIFAR}\label{sec:pl}
To evaluate the performance of our method in the settings of transfer learning, we trained the network on PL-CIFAR by using our method. In the first step of our method, we used a learning rate of 0.04 and used 1.2 for $\alpha$ and 0.8 for $\beta$. As a comparison, we also trained on initial pseudo-labels in the same manner as the second step of our method.

\begin{figure}[tb]
  \vspace{-5mm}
  \centering
  \scalebox{0.5}{
  \includegraphics{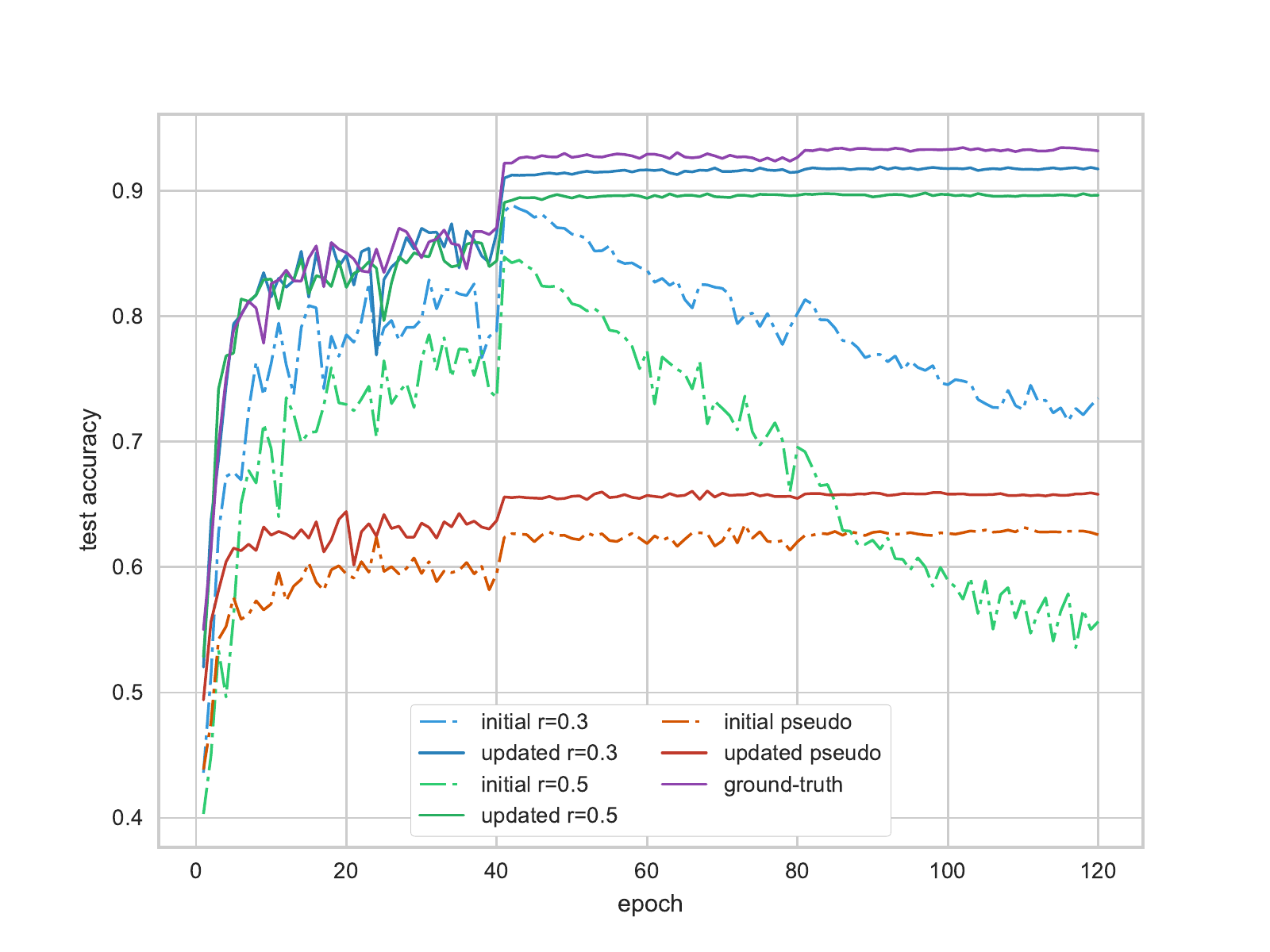}
  }
  \caption{The test accuracy curve with different labels. The test accuracy on the pseudo labels is lower than that on the symmetric noise labels even if the number of inaccurate labels is lower. This trend remains if the labels are updated.}
  \label{fig:trans_a}
  \vspace{-4mm}
\end{figure}

\begin{figure}[tb]
  \vspace{-5mm}
  \centering
  \scalebox{0.5}{
  \includegraphics{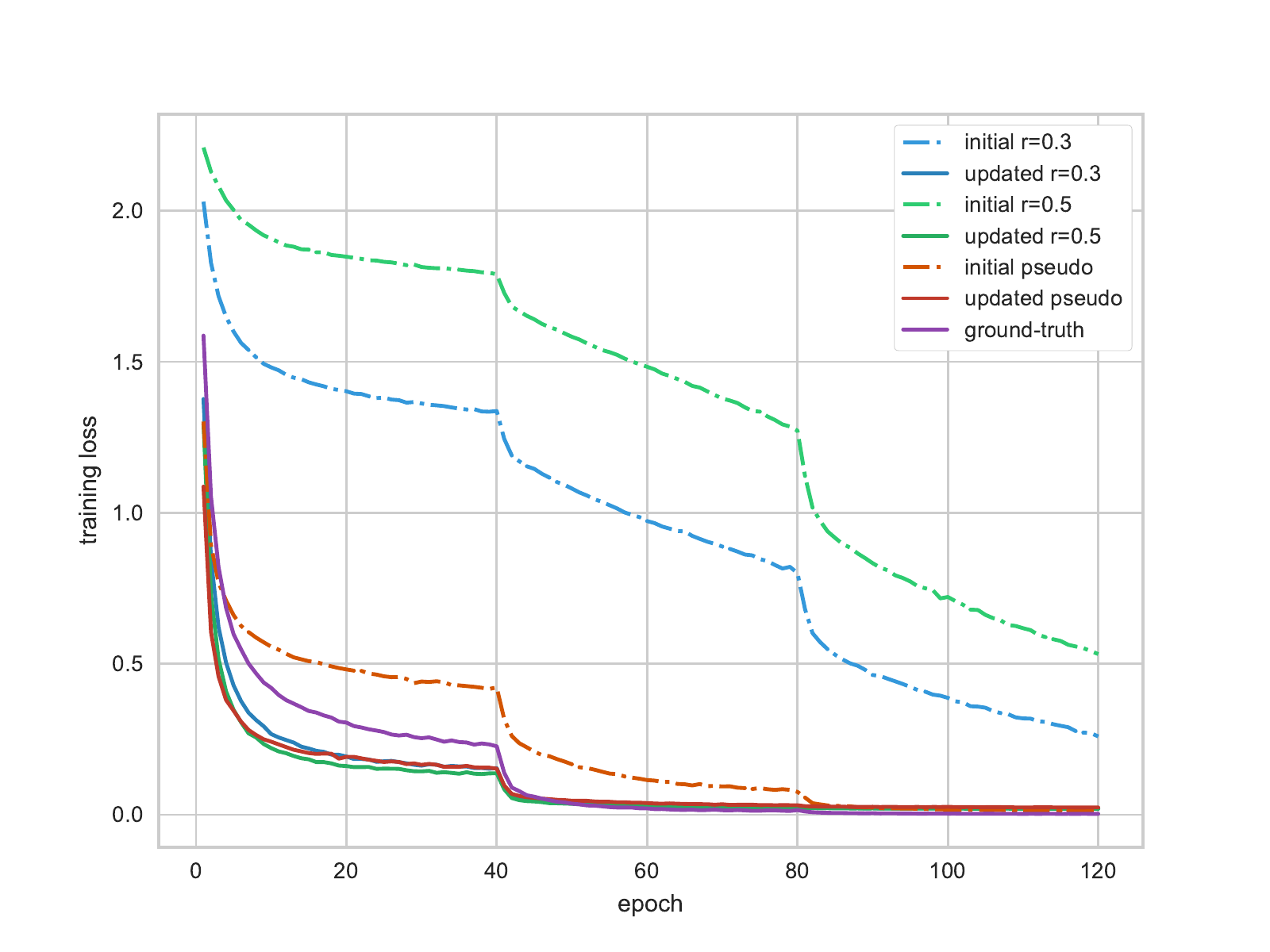}
  }
  \caption{The training loss curve with different labels. The training loss on the pseudo labels is lower than that on the symmetric noise labels even if the number of inaccurate labels is higher. Each of the training losses on the labels updated from different noisy labels follows the training loss on the ground-truth labels. This implies that the updated labels are completely optimized for the network.}
  \label{fig:trans_l}
\end{figure}

\Fref{fig:trans_a} shows the test accuracy curve with different labels, and \Fref{fig:trans_l} shows the decline in the training loss during training. In both figures, we show the results of training on SN-CIFAR (the noise rate $r=0.3,0.5$) because the noise rate of the pseudo labels is between 0.3 and 0.5. Additionally, we show the results of training on the ground-truth labels because the training loss curve of training on optimized labels is near the curve for the same.

Although the number of inaccurate labels in the pseudo labels exceeds that of the symmetric noise labels ($r=0.3$), the value of the training loss of the pseudo labels is lower than that of the symmetric noise labels. This fact seems to conflict with extant knowledge that states that ``the training loss increases when the noise rate increases'', as described in \Sref{sec:prelim}. However, we can explain the reason of this conflict as follows: the difference in the training loss depends on the noise rate as well as the type of the noise.
The pseudo labels are generated from the outputs of ResNet-50 pre-trained on ImageNet, and thus they are already considered as ``the optimized labels'' by the network. Thus, the pseudo labels were not updated adequately.
The test accuracy of training on the labels recovered from the noisy labels is worse than that of training on the ground-truth labels, and this indicates that the optimized labels do not necessarily denote optimal labels. This is a limitation of the proposed method.

\subsection{Experiment on the Clothing1M dataset}\label{sec:cloth}

Finally, we trained the network on the Clothing1M dataset~\cite{xiao2015learning} by using our method to evaluate the performance of our method in a real setting. As a comparison, we also trained on initial noisy labels in the same manner as the second step of our method. The results of our experiments are shown in \Tref{tab:cloth}. Additionally, we also show the scores (\#1, \#2) reported in~\cite{patrini2016making}.

In \#2, Patrini \etal exploited the curated labels of $50k$ clean data and their noisy versions in $1M$ noisy data to obtain the ground-truth noise transition matrix,
which is not often used in real-world settings. Conversely, we only used the distribution of the noisy labels, which can be always used, for the prior distribution $p$,
and therefore our condition is more general than \#2.
Nevertheless, our method achieves better test accuracy than \#2 on the Clothing1M dataset.

\begin{figure}[tb]
  \centering
  \scalebox{0.38}{
  \includegraphics{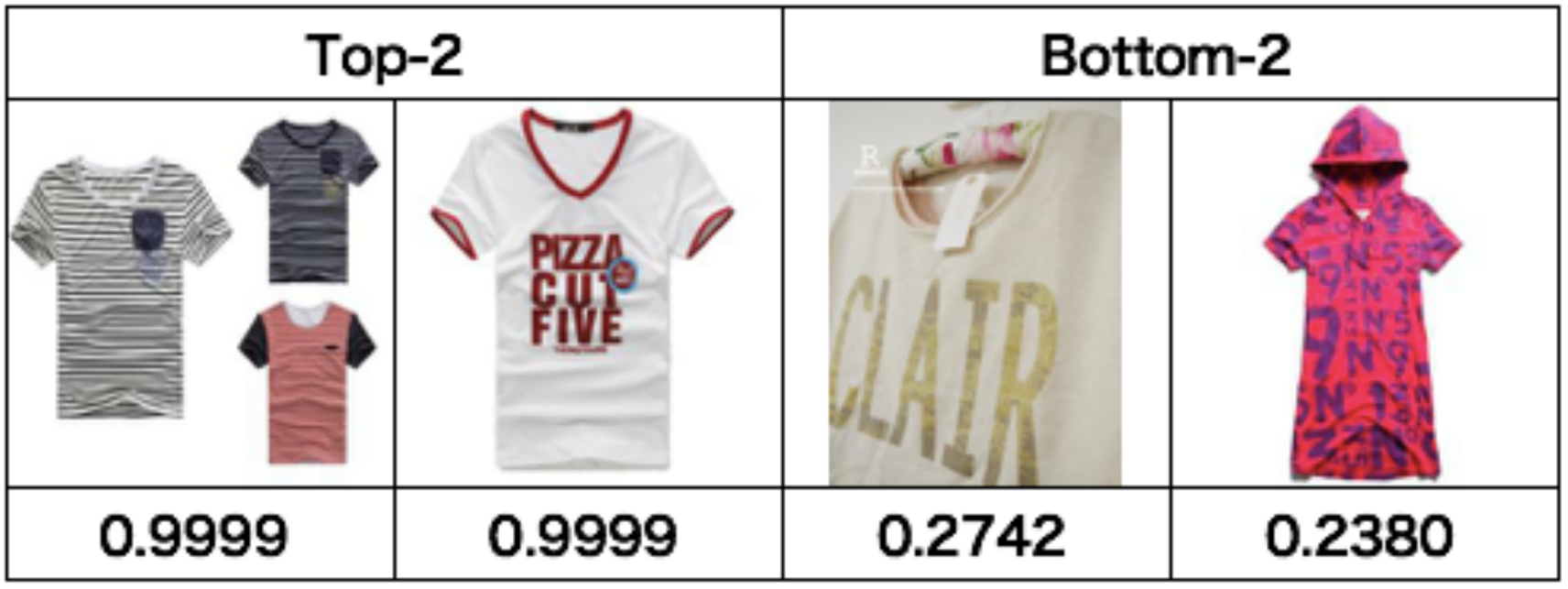}
  }
 \caption{The images with the top-2 and the bottom-2 probabilities of T-shirt whose labels are reassigned from Hoodie to T-shirt.}
 \label{fig:H2T}
 \vspace{-1mm}
\end{figure}

\begin{figure}[tb]
  \centering
  \scalebox{0.38}{
  \includegraphics{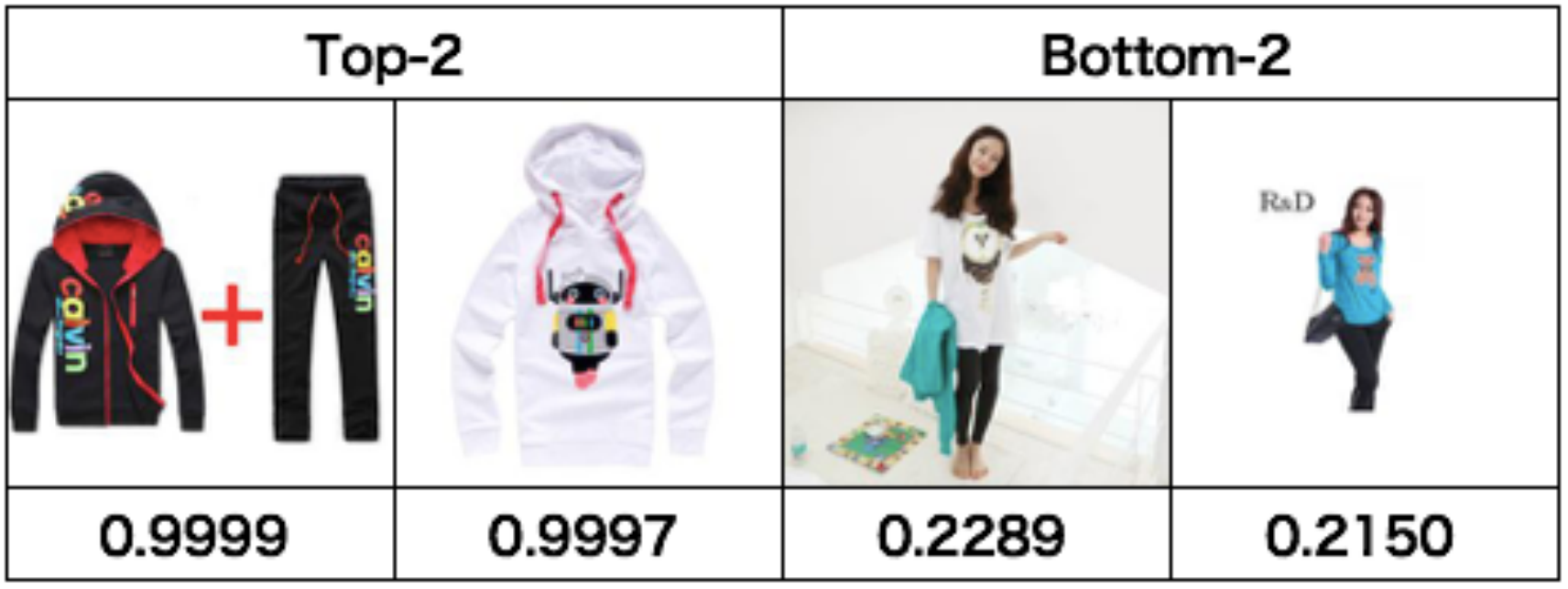}
  }
 \caption{The images with the top-2 and the bottom-2 probabilities of Hoodie whose labels are reassigned from T-shirt to Hoodie.}
 \label{fig:T2H}
 \vspace{-5mm}
\end{figure}

In \Fref{fig:H2T}, \Fref{fig:T2H}, we show the examples of the images whose labels were reassigned to classes different from the original ones by our method. Additionally, we show the probability of the class that the label is reassigned to.
When the probability is high, the label seems to be updated correctly. Conversely, when the probability is low, the label seems to be updated incorrectly. As opposed to the hard-labels, the soft-labels contain the probabilities of each class in themselves, and thus the network can consider the incorrectly updated labels as not important. Specifically, this effect contributes to improving the test accuracy.

\begin{table}[tb]
  \caption{Test accuracy of different baselines on the Clothing1M dataset. \#1 and \#2 are quoted from~\cite{patrini2016making}, and \#3 is the reproduced result by our reimplementation.}
  \vspace{-2mm}
  \begin{center}
  \scalebox{0.9}{
  \begin{tabular}{c|cc|c}\hline
    \# & method &  & accuracy\\ \hline\hline
    1 & Cross Entropy Loss & & 68.94\\ \hline
    2 & Forward~\cite{patrini2016making} & & 69.84\\ \hline
    3 & Cross Entropy Loss & \textit{best} & 69.15\\
    &(reproduced) & \textit{last} & 66.76\\ \hline
    4& Our Method & \textit{best} & 72.16\\
    && \textit{last} & \textbf{72.23}\\ \hline
  \end{tabular}
  }
  \end{center}
  \label{tab:cloth}
  \vspace{-3mm}
\end{table}

\section{Conclusion}
We proposed a joint optimization framework for learning on noisy labeled datasets, which alternatively updates network parameters and class labels. The performance of the framework is guaranteed by our finding that training under a high learning rate prevents the network from memorizing noisy labels. We showed that our framework performed remarkably well on the noisy CIFAR-10 dataset and the Clothing1M dataset, outperforming the state-of-the-art methods~\cite{patrini2016making,vahdat2017toward}.

\vspace{2mm}\noindent\textbf{Acknowledgements. }
This research is partially supported by CREST (JPMJCR1686).

\newpage
{\small
\bibliographystyle{ieee}
\bibliography{references}
}

\newpage
\appendix
{\Huge{\bf Appendices}}
\section{Detailed Architecture}\label{sec:arch}
\Tref{tab:arch} details the network architecture used in the experiments on the CIFAR-10 dataset. It is based on PreAct ResNet-32~\cite{he2016identity}.

\begin{table}[h]
  \caption{The network architecture used in the experiments on CIFAR-10.}
  \begin{center}
  \scalebox{0.9}{
  \begin{tabular}{l|l}\hline
    NAME & DESCRIPTION \\ \hline\hline
    input & 32$\times$32 RGB imgae \\ \hline
    conv & 32 filters, 3$\times$3, pad=1, stride=1 \\ \hline
    unit1 & (pre-activation Residual Unit 32$\rightarrow$32)$\times$5 \\ \hline
    unit2a & pre-activation Residual Unit 32$\rightarrow$64 \\ \hline
    unit2b & (pre-activation Residual Unit 64$\rightarrow$64)$\times$4 \\ \hline
    unit3a & pre-activation Residual Unit 64$\rightarrow$128 \\ \hline
    unit3b & (pre-activation Residual Unit 128$\rightarrow$128)$\times$4 \\ \hline
    pool & Batch Normarization, ReLU, \\
     & Global average pool (8$\times$8$\rightarrow$1$\times$1 pixels)\\ \hline
    dense & Fully connected 128$\rightarrow$10 \\ \hline
    output & Softmax \\ \hline
  \end{tabular}
  }
  \end{center}
  \label{tab:arch}
\end{table}

\section{Dependency on Hyper Parameters}\label{sec:hyper}
We show the hyper parameters used in the experiments on SN-CIFAR in \Tref{tab:snhyper}. If the noise rate is high, the optimal learning rate also tends to be high.

\begin{table}[h]
  \caption{The hyper parameters used in the experiments on SN-CIFAR.}
  \vspace{2mm}
  \centering
  \scalebox{0.75}{
  \begin{tabular}{c|cccccc}\hline
    \multicolumn{1}{c|}{noise rate (\%)}&0&10&30&50&70&90\\ \hline\hline
    \multicolumn{1}{c|}{$\alpha$}& 1.2 & 1.2 & 1.2 & 1.2 & 1.2 & 0.8\\ \hline
    \multicolumn{1}{c|}{$\beta$}& 0.8 & 0.8 & 0.8 & 0.8 & 0.8 & 0.4\\ \hline
    \multicolumn{1}{c|}{learning rate}& 0.01 & 0.02 & 0.03 & 0.04 & 0.08 & 0.12\\ \hline
  \end{tabular}
  }
  \label{tab:snhyper}
\end{table}

The prediction accuracy is not so sensitive to the hyper parameters and our method demonstrated good performance with a different set of the hyper parameters as shown in \Tref{tab:hyper1}, \ref{tab:hyper2}, \ref{tab:hyper3}, \ref{tab:hyper4}.
In addition, \Tref{tab:begin1},~\ref{tab:begin2} show the validation accuracy with different $t_1$ and $t_2$, where $t_1$ is the value at which to start label-updating, and $t_2$ is the value at which to stop label-updating.
When we train the network with a high learning rate, the prediction accuracy retains high value, and thus we can start label-updating when the validation accuracy once reach high value. Label-updating should be stopped after the training loss converge.

\begin{table}[b]
  \vspace{-5pt}
  \caption{Validation accuracy with different hyper parameters in the triple test (experimented on AN-CIFAR with noise rate $=0.4$).}
  \centering
  \scalebox{0.75}{
  \begin{tabular}{ccc|ccccccc}\hline
    \multicolumn{10}{c}{$\beta=0.4$, learning rate $=0.03$}\\ \hline
    \multicolumn{3}{c|}{$\alpha$}&0.1&0.2&0.5&{\bf 0.8}&1.0&2.0&5.0\\ \hline\hline
    \multicolumn{3}{c|}{val (\%)}& 91.9 & 92.0 & 91.7 & 92.0 & 92.1 & 92.1 & 88.8\\ \hline
    \multicolumn{10}{c}{}\\ \hline
    \multicolumn{10}{c}{$\alpha=0.8$, learning rate $=0.03$}\\ \hline
    \multicolumn{3}{c|}{$\beta$}&0.05&0.1&0.2&{\bf 0.4}&0.5&1.0&2.0\\ \hline\hline
    \multicolumn{3}{c|}{val (\%)}& 90.8 & 91.7 & 91.8 & 92.0 & 91.6 & 89.5 & 91.1\\ \hline
    \multicolumn{10}{c}{}\\ \hline
    \multicolumn{10}{c}{$\alpha=0.8$, $\beta=0.4$}\\ \hline
    \multicolumn{3}{c|}{learning rate}&0.005&0.01&0.02&{\bf 0.03}&0.05&0.1&0.2\\ \hline\hline
    \multicolumn{3}{c|}{val (\%)}& 90.6 & 90.9 & 91.3 & 92.0 & 92.1 & 91.3 & 88.5\\ \hline
  \end{tabular}
  }
  \label{tab:hyper1}
\end{table}

\begin{table}[b]
  \caption{Validation accuracy with different hyper parameters in the triple test (experimented on AN-CIFAR with noise rate $=0.2$).}
  \centering
  \scalebox{0.75}{
  \begin{tabular}{ccc|ccccccc}\hline
    \multicolumn{10}{c}{$\beta=0.4$, learning rate $=0.03$}\\ \hline
    \multicolumn{3}{c|}{$\alpha$}&0.1&0.2&0.5&{\bf 0.8}&1.0&2.0&5.0\\ \hline\hline
    \multicolumn{3}{c|}{val (\%)}& 92.9 & 92.9 & 93.0 & 93.2 & 93.1 & 93.2 & 89.7\\ \hline
    \multicolumn{10}{c}{}\\ \hline
    \multicolumn{10}{c}{$\alpha=0.8$, learning rate $=0.03$}\\ \hline
    \multicolumn{3}{c|}{$\beta$}&0.05&0.1&0.2&{\bf 0.4}&0.5&1.0&2.0\\ \hline\hline
    \multicolumn{3}{c|}{val (\%)}& 92.6 & 93.0 & 93.2 & 93.2 & 93.1 & 92.8 & 92.8\\ \hline
    \multicolumn{10}{c}{}\\ \hline
    \multicolumn{10}{c}{$\alpha=0.8$, $\beta=0.4$}\\ \hline
    \multicolumn{3}{c|}{learning rate}&0.005&0.01&0.02&{\bf 0.03}&0.05&0.1&0.2\\ \hline\hline
    \multicolumn{3}{c|}{val (\%)}& 92.5 & 92.7 & 92.7 & 93.2 & 92.7 & 91.8 & 89.2\\ \hline
  \end{tabular}
  }
  \label{tab:hyper2}
\end{table}

\begin{table}[b]
  \caption{Validation accuracy with different hyper parameters in the triple test (experimented on SN-CIFAR with noise rate $=0.7$).}
  \centering
  \scalebox{0.75}{
  \begin{tabular}{ccc|ccccccc}\hline
    \multicolumn{10}{c}{$\beta=0.8$, learning rate $=0.08$}\\ \hline
    \multicolumn{3}{c|}{$\alpha$}&0.1&0.2&0.5&1.0&{\bf 1.2}&2.0&5.0\\ \hline\hline
    \multicolumn{3}{c|}{val (\%)}& 85.7 & 86.0 & 85.5 & 85.9 & 85.5 & 85.7 & 83.8\\ \hline
    \multicolumn{10}{c}{}\\ \hline
    \multicolumn{10}{c}{$\alpha=1.2$, learning rate $=0.08$}\\ \hline
    \multicolumn{3}{c|}{$\beta$}&0.05&0.1&0.2&0.5&{\bf 0.8}&1.0&2.0\\ \hline\hline
    \multicolumn{3}{c|}{val (\%)}& 82.0 & 82.3 & 83.1 & 85.3 & 85.5 & 85.2 & 30.3\\ \hline
    \multicolumn{10}{c}{}\\ \hline
    \multicolumn{10}{c}{$\alpha=1.2$, $\beta=0.8$}\\ \hline
    \multicolumn{3}{c|}{learning rate}&0.005&0.01&0.02&0.05&{\bf 0.08}&0.1&0.2\\ \hline\hline
    \multicolumn{3}{c|}{val (\%)}& 79.5 & 80.7 & 82.8 & 85.4 & 85.5 & 85.4 & 83.8\\ \hline
  \end{tabular}
  }
  \label{tab:hyper3}
\end{table}

\begin{table}[bt]
  \caption{Validation accuracy with different hyper parameters in the triple test (experimented on SN-CIFAR with noise rate $=0.3$).}
  \centering
  \scalebox{0.75}{
  \begin{tabular}{ccc|ccccccc}\hline
    \multicolumn{10}{c}{$\beta=0.8$, learning rate $=0.03$}\\ \hline
    \multicolumn{3}{c|}{$\alpha$}&0.1&0.2&0.5&1.0&{\bf 1.2}&2.0&5.0\\ \hline\hline
    \multicolumn{3}{c|}{val (\%)}& 91.6 & 91.7 & 91.5 & 91.8 & 91.8 & 91.8 & 89.9\\ \hline
    \multicolumn{10}{c}{}\\ \hline
    \multicolumn{10}{c}{$\alpha=1.2$, learning rate $=0.03$}\\ \hline
    \multicolumn{3}{c|}{$\beta$}&0.05&0.1&0.2&0.5&{\bf 0.8}&1.0&2.0\\ \hline\hline
    \multicolumn{3}{c|}{val (\%)}& 90.0 & 90.4 & 91.2 & 91.8 & 91.8 & 91.9 & 91.0\\ \hline
    \multicolumn{10}{c}{}\\ \hline
    \multicolumn{10}{c}{$\alpha=1.2$, $\beta=0.8$}\\ \hline
    \multicolumn{3}{c|}{learning rate}&0.005&0.01&0.02&{\bf 0.03}&0.05&0.1&0.2\\ \hline\hline
    \multicolumn{3}{c|}{val (\%)}& 90.1 & 90.7 & 91.0 & 91.8 & 92.1 & 91.1 & 89.0\\ \hline
  \end{tabular}
  }
  \label{tab:hyper4}
\end{table}

\begin{table}[bt]
  \caption{Validation accuracy with different $t_1$ (start epoch) and $t_2$ (stop epoch) in the triple test (experimented on AN-CIFAR with noise rate $=0.4$, $\alpha=0.8$, $\beta=0.4$, learning rate = 0.1).}
  \centering
  \scalebox{0.9}{
  \begin{tabular}{ccc|ccccc}\hline
    \multicolumn{3}{c|}{start epoch}&0&50&{\bf 70}&100&150\\ \hline\hline
    \multicolumn{3}{c|}{val (\%)}& 58.4 & 90.3 & 91.3 & 91.4 & 91.6\\ \hline
    \multicolumn{8}{c}{}\\ \hline
    \multicolumn{3}{c|}{stop epoch}&100&150&{\bf 200}&250&300\\ \hline\hline
    \multicolumn{3}{c|}{val (\%)}& 91.8 & 91.5 & 91.3 & 90.8 & 90.7\\ \hline
  \end{tabular}
  }
  \label{tab:begin1}
\end{table}

\begin{table}[bt]
  \caption{Validation accuracy with different $t_1$ (start epoch) and $t_2$ (stop epoch) in the triple test (experimented on SN-CIFAR with noise rate $=0.7$, $\alpha=1.2$, $\beta=0.8$, learning rate = 0.08).}
  \centering
  \scalebox{0.9}{
  \begin{tabular}{ccc|ccccc}\hline
    \multicolumn{3}{c|}{start epoch}&0&50&{\bf 70}&100&150\\ \hline\hline
    \multicolumn{3}{c|}{val (\%)}& 38.0 & 84.7 & 85.5 & 86.1 & 85.6\\ \hline
    \multicolumn{8}{c}{}\\ \hline
    \multicolumn{3}{c|}{stop epoch}&100&150&{\bf 200}&250&300\\ \hline\hline
    \multicolumn{3}{c|}{val (\%)}& 85.0 & 85.6 & 85.5 & 85.9 & 85.6\\ \hline
  \end{tabular}
  }
  \label{tab:begin2}
\end{table}

\newpage
\section{Effect of Soft-Labeling}\label{sec:effect}
We show the analysis of the effect of soft-labeling on the noisy CIFAR-10 dataset in \Tref{tab:rec1},~\ref{tab:rec2}.
The soft-labels with high probability are almost correct.
Conversely, when the probability is low, the label seems to be updated incorrectly. As opposed to the hard-labels, the soft-labels contain the probabilities of each class in themselves, and thus the network can consider the incorrectly updated labels as not important.

\begin{table}[h]
  \caption{Recovery accuracies of the updated soft-labels whose maximum probabilities $p$ are within each range (experimented on AN-CIFAR with noise rate $=0.4$).}
  \centering
  \scalebox{0.75}{
  \begin{tabular}{ccc|cccc|c}\hline
    \multicolumn{3}{c|}{$p$}&$1-0.99$&$0.99-0.95$&$0.95-0.9$&$0.9-0$&$1-0$\\ \hline\hline
    \multicolumn{3}{c|}{acc (\%)}& 99.8 & 96.9 & 91.3 & 73.1 & 95.1\\ \hline
    \multicolumn{3}{c|}{number}& 27046 & 8647 & 3484 & 5823 & 45000\\ \hline
  \end{tabular}
  }
  \label{tab:rec1}
\end{table}

\begin{table}[h]
  \caption{Recovery accuracies of the updated soft-labels whose maximum probabilities $p$ are within each range (experimented on SN-CIFAR with noise rate $=0.7$).}
  \centering
  \scalebox{0.75}{
  \begin{tabular}{ccc|cccc|c}\hline
    \multicolumn{3}{c|}{$p$}&$1-0.99$&$0.99-0.95$&$0.95-0.9$&$0.9-0$&$1-0$\\ \hline\hline
    \multicolumn{3}{c|}{acc (\%)}& 97.5 & 82.2 & 70.6 & 53.3 & 86.4\\ \hline
    \multicolumn{3}{c|}{number}& 27591 & 7368 & 3351 & 6690 & 45000\\ \hline
  \end{tabular}
  }
  \label{tab:rec2}
\end{table}

\end{document}